\documentclass[final]{cvpr}

\usepackage[utf8]{inputenc}
\usepackage{times}
\usepackage{epsfig}
\usepackage{graphicx}
\usepackage{amsmath}
\usepackage{amssymb}

\usepackage{amsfonts}
\usepackage{makecell}

\newcommand*\samethanks[1][\value{footnote}]{\footnotemark[#1]}

\usepackage{cite} 
\usepackage{microtype} 

\usepackage{tabularx}
\usepackage{booktabs}
\usepackage{array}
\newcolumntype{H}{>{\setbox0=\hbox\bgroup}c<{\egroup}@{}}

\usepackage{mathtools}

\usepackage{xcolor}
\usepackage{url}
\usepackage[pagebackref=true,breaklinks=true,colorlinks,bookmarks=false]{hyperref}

\title{Recurrently Estimating Reflective Symmetry Planes from Partial Pointclouds}

\author{
  Mihaela C\u{a}t\u{a}lina Stoian\thanks{\{mihaela.stoian,tommaso.cavallari\}@five.ai} \qquad Tommaso Cavallari\samethanks\\
  Five \\
  Oxford, UK  
}

\def\cvprPaperID{10}
\def\confYear{3DVR 2021}

\begin{document}
\maketitle

\begin{abstract}
Many man-made objects are characterised by a shape that is symmetric along one or more planar directions.
Estimating the location and orientation of such symmetry planes can aid many tasks such as estimating the overall orientation of an object of interest or performing shape completion, where a partial scan of an object is reflected across the estimated symmetry plane in order to obtain a more detailed shape.
Many methods processing 3D data rely on expensive 3D convolutions.
In this paper we present an alternative novel encoding that instead slices the data along the height dimension and passes it sequentially to a 2D convolutional recurrent regression scheme.
The method also comprises a differentiable least squares step, allowing for end-to-end accurate and fast processing of both full and partial scans of symmetric objects.
We use this approach to efficiently handle 3D inputs to design a method to estimate planar reflective symmetries.
We show that our approach has an accuracy comparable to state-of-the-art techniques on the task of planar reflective symmetry estimation on full synthetic objects.
Additionally, we show that it can be deployed on partial scans of objects in a real-world pipeline to improve the outputs of a 3D object detector.
\end{abstract}

%%%%%%%%%%%%%%%%%%%%%%%%%%%%%%%%%%%%%%%%%%%%%%%%%%%%%%%%%%%%%%%%%%%%%%%%%%%%%%%%%%%%%%%%%%%%%%%%%%%%%%%%%%%%%%%%%%%%%%%%%%%%%%%%%%%%%%%%%%
\section{Introduction}
Processing 3D objects has proved useful (and, many times, necessary) in many scenarios, including robot grasping, autonomous driving, navigation and mapping. 
Systems deployed within these fields often encounter many sub-problems that have to be tackled independently, but whose solution is crucial to the behaviour of the whole system.
Such problems include object detection, pose estimation, shape completion and matching, and many others.
Symmetry detection has a high potential for aiding each of these tasks.
While commonly investigated in 2D scenarios~\cite{Zielke1992, Chen2018, Funk2017a, Cicconet2017, Elawady2017, Vasudevan2018}, it gained more attention recently in 3D vision~\cite{Bohg2011, Ilonen2014, Gao2017, Ecins2017, Funk2017, Cicconet2017b, Ecins2018, Makhal2018, Gao2020, Shi2020}.

Found in natural and man-made structures, symmetry appears locally and often also globally within objects. 
For the purposes of this paper, we focus on global symmetry detection from 3D objects and consider an object symmetric if there is at least one plane such that by reflecting the object across the plane we get (approximately) the same object; otherwise, we say that the object is asymmetric.

For most tasks dealing with 3D objects, one of the challenges is choosing how to represent the input data.
Options include: handling meshes or pointclouds directly (e.g.~\cite{Ecins2017, Makhal2018}), discretising them into voxels and performing 3D convolutions (e.g.~\cite{Gao2020}), or relying on 2D representations (e.g. RGB-D images) and using standard network architectures to process them (e.g.~\cite{Shi2020}).
In this paper we present a novel approach which combines a voxelisation of the input with an efficient recurrent processing based on 2D convolutions performed sequentially over different horizontal slices of the data.
Differently from using expensive 3D convolutions over the whole voxelised input, this allows us to use a larger resolution, thus capturing more details and leading to a more robust system.
Slicing the data and processing each chunk sequentially is advantageous as the network can share parameters across slices and still attend to fine details in the input.
Notably, this novel way of processing the input is not specific to symmetry detection and can be easily deployed in any setting requiring convolutional processing of 3D data.

\begin{figure*}[ht!]
  \centering
  \includegraphics[width=0.9\textwidth,height=3.5cm]{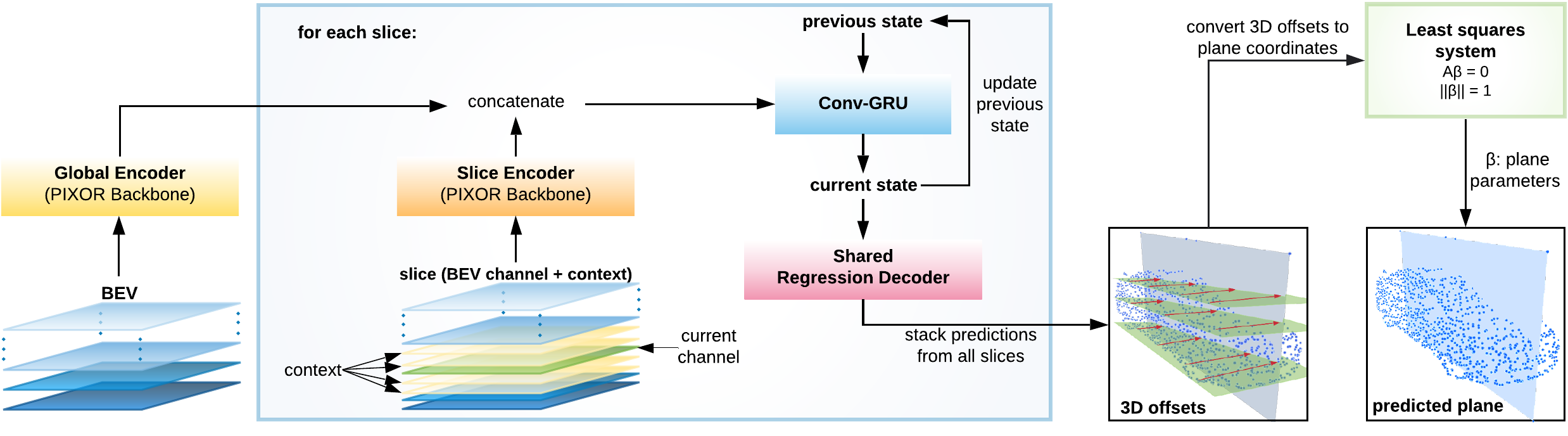} % 
  \caption{Overview of the pipeline we designed to estimate symmetry planes from 3D pointclouds.}
  \label{fig:architecture-overview}
  \vspace*{-0.5cm}
\end{figure*}

We equip our model with a convolutional gated recurrent unit (ConvGRU) network to help it predict planes when the objects are arbitrarily rotated in 3D space (as the symmetry planes would intersect with each slice in a different location).
The ConvGRU observes different slices of the input pointcloud sequentially, considering each slice as a time step.
For each slice, the network regresses per-pixel locations of the symmetry plane relative to the slice itself.
These predictions then allow us to estimate the plane by minimising a differentiable least-squares system (inspired by~\cite{Gansbeke2019}).

As currently there is no standard dataset for the planar reflective symmetry estimation task, we report the performance of our approach on the benchmark recently introduced by the \textit{PRS-Net} paper~\cite{Gao2020} and based on ShapeNet~\cite{Chang2015}.
We also show that our method can be applied to self-obstructed or partially occluded objects, such as those captured by a sensor (e.g. LiDAR) in the real-world.
In this scenario, we use the detected symmetry planes to refine bounding boxes output by a 3D object detector, and empirically determine that it is possible to reduce the angular error of such boxes.
As mentioned, the object detection task is a crucial component that is often one of the first steps of many complex pipelines in robotic navigation or grasping, autonomous driving, unsupervised or automatic labelling/annotation.
Therefore, improving the accuracy of the detections can significantly increase the chance of success of each pipeline.

\noindent To summarise, the main contributions of this paper are: 
\vspace*{-0.25cm}
\begin{itemize}
\setlength\itemsep{-0.3em}
    \item A novel way to process 3D data represented by a voxel grid via slicing its contents along the height dimension;
    \item A network based on ConvGRU which regresses 3D offsets indicating where a reflective symmetry plane would lie relatively to a regularly sampled grid of points on the object or partial scan;
    \item A differentiable constrained least-squares solver which is used to estimate the parameters of a plane passing through the points determined by the 3D offsets predicted by the ConvGRU network;
    \item A method that uses the estimated planes to refine boxes output by a 3D object detector, improving its accuracy.
\end{itemize}

%%%%%%%%%%%%%%%%%%%%%%%%%%%%%%%%%%%%%%%%%%%%%%%%%%%%%%%%%%%%%%%%%%%%%%%%%%%%%%%%%%%%%%%%%%%%%%%%%%%%%%%%%%%%%%%%%%%%%%%%%%%%%%%%%%%%%%%%%%
\section{Method}
Our method performs symmetry detection for pointclouds depicting symmetric 3D objects: either full or partially occluded objects from ShapeNet, or real-world data consisting of partial clouds such as those gathered from LiDAR scans.

The input pointclouds are voxelised to a top-down Bird's Eye View (BEV) representation, which is then partitioned along its height dimension (not necessarily corresponding to any canonical vertical direction of the objects) to form slices comprising multiple contiguous channels. 
Our architecture consists of two encoders (global -- observing the whole BEV representation -- and per-slice), a ConvGRU component, and a decoder (regressing per-pixel 3D offsets indicating where the symmetry plane lies relatively to each slice pixel).
These are followed by a least-squares regression module, as shown in \autoref{fig:architecture-overview}.

In what follows we will denote symmetry planes by $\mathcal{S} = (\mathbf{n},d)$, where $\mathbf{n}=(a,b,c)$ is a unit vector defining the normal of the plane, $d$ is the plane's distance from the origin, and  $ax + by + cz = d$ for every point $\mathbf{p} = (x, y, z)$ lying on the plane $\mathcal{S}$.
\autoref{eq:refl} describes the reflection of a general point $\mathbf{p} \in \mathbb{R}^3$ across a symmetry plane $\mathcal{S}$.
\begin{align} \label{eq:refl}
    R_\mathcal{S}(\mathbf{p}) &= \mathbf{p} - 2\mathbf{n}( \mathbf{p} \cdot \mathbf{n} - d)
\end{align}

%%%%%%%%%%%%%%%%%%%%%%%%%%%%%%%%%%%%%%%%%%%%%%%%%%%%%%%%%%%%%%%%%%%%%%%%%%%%%%%%%%%%%%%%%%%%%%%%%%%%%%%%%%%%%%%%%%%%%%%%%%%%%%%%%%%%%%%%%%
\subsection{Encoding}
\label{sec:bev}
We use a top-down projection to discretise sparse unordered 3D pointclouds into dense regular BEV representations.
We simply create an axis-aligned binary occupancy grid (of size $H \times D \times W$) by voxelising the space contained within a unit box centred on the origin.
We ensure that the points fed as input fit in the unit box by normalising their coordinates.
The BEV representation is best suited to be processed by 2D convolutions over ``pixels'' located on a plane parallel to the ground (horizontal) plane.
Nevertheless, we do not lose the information coming from the third dimension as we consider all the voxels located above each ground-plane pixel as $H$ separate channels that are fed as input to the 2D convolutions.
All these channels are passed to the \textbf{global} encoder at once.

Additionally, we \textit{slice} the BEV representation in order to obtain multiple views of limited regions of the 3D space.
More specifically, we choose $N$ channels of interest over the height dimension of the voxelisation and, given a size $K$, for each of these we select as context the previous and subsequent $K$ channels to form a corresponding slice.
These slices are fed sequentially to the \textbf{slice} encoder, starting from the bottom of the BEV representation (see~\autoref{fig:architecture-overview}, bottom-centre).
For each of the $N$ slices, the output of the slice encoder is concatenated to the output of the global encoder.

Both encoders are based on PIXOR's~\cite{Yang2018} backbone, which performs 2D convolutions over the height channels of the BEV representation to output a high-resolution feature map four times smaller than the input.
We empirically found that the global and slice encoders are complementary, as it was not enough to use any of the two by themselves.

%%%%%%%%%%%%%%%%%%%%%%%%%%%%%%%%%%%%%%%%%%%%%%%%%%%%%%%%%%%%%%%%%%%%%%%%%%%%%%%%%%%%%%%%%%%%%%%%%%%%%%%%%%%%%%%%%%%%%%%%%%%%%%%%%%%%%%%%%%
\subsection{ConvGRU} 
\label{sec:convgru}
As we process the chunks of BEV channels \textit{sequentially}, we want to compute the predictions at each slice keeping into account the information we gained from processing the previous slices as well.
Thus, we use a \textit{recurrent} approach.
We choose Gated Recurrent Units (GRUs)~\cite{Cho2014} as they are compact, but comparable with other RNN-based models such as LSTMs in terms of performance~\cite{Chung2014}.
To account for the higher dimensionality of our input to the GRUs -- sequences of 2D feature maps instead of 1D sequences used traditionally -- we change the update and reset gates to perform 2D convolutions rather than dot products, similarly to \cite{Choy2016, Siam2017}, obtaining a Convolutional GRU (ConvGRU).

\autoref{fig:architecture-overview} shows a simplified version of the architecture using a single ConvGRU layer (in practice we use three).
We initialise each hidden state with a separate 3x3 convolutional layer -- processing the output of the global encoder -- followed by group normalization (GN)~\cite{Wu2018} and ReLU.
The hidden state output by the last ConvGRU layer for each time step (i.e. slice) is passed to a decoder, shared between the slices, that predicts the corresponding 3D offsets.

%%%%%%%%%%%%%%%%%%%%%%%%%%%%%%%%%%%%%%%%%%%%%%%%%%%%%%%%%%%%%%%%%%%%%%%%%%%%%%%%%%%%%%%%%%%%%%%%%%%%%%%%%%%%%%%%%%%%%%%%%%%%%%%%%%%%%%%%%%
\subsection{Decoding and estimating symmetry planes}
\label{sec:decoder}
The decoder consists of five 3x3 convolutional layers, where the first four are followed by GN and ReLU, and the last one yields 3D offsets.
Each 3D offset indicates the relative location of the symmetry plane w.r.t. each pixel element in the slice.
After all the slices have been processed, the offsets are stacked into a single tensor.
As the offsets predicted by the decoder are relative to the corresponding pixels from each slice, we convert them to world-space coordinates representing a set of points $P$ lying on the symmetry plane.

In order to find the optimal symmetry plane $\hat{\mathcal{S}} = (\hat{\mathbf{n}},\hat{d})$ passing through these points we solve a least squares system of the form $A \mathbf{\beta} = \mathbf{0}$, where $A$ is a $N \times 4$ matrix obtained by stacking the homogeneous representation of the points in $P$, and $\beta$ are the parameters of the plane $\hat{\mathcal{S}}$. 
As we want to avoid obtaining the trivial solution of $\mathbf{\beta} = \mathbf{0}$, we also introduce a constraint on the norm of the parameters of the system, $|| \mathbf{\beta} || = 1$, and solve it using the technique from Appendix~A5 in \cite{Hartley2004}.
It is worthwhile noting that estimating the parameters of the plane remains a fully differentiable operation.
Thus, we can backpropagate the losses through it.

%%%%%%%%%%%%%%%%%%%%%%%%%%%%%%%%%%%%%%%%%%%%%%%%%%%%%%%%%%%%%%%%%%%%%%%%%%%%%%%%%%%%%%%%%%%%%%%%%%%%%%%%%%%%%%%%%%%%%%%%%%%%%%%%%%%%%%%%%%
\subsection{Losses}
\label{sec:losses}
For supervision we use the ground truth symmetry planes to compute two losses: one based on the offsets predicted by the model and another one based on the plane parameters. 
For evaluation, we also use a geometric loss to measure the quality of the predicted plane.

\vspace*{-0.3cm}
\paragraph*{Offsets Loss.}
Given the world-space coordinates $\mathbf{p}$ of the center of each voxel in the $N$ channels selected when slicing the BEV representation, we can easily compute their relative offset $\mathbf{y}$ from the ground truth symmetry plane $\mathcal{S} = (\mathbf{n}, d)$ using the signed point-to-plane distance:
\vskip-0.3cm
\begin{equation}
    \mathbf{y} = -(\mathbf{n p} - d)\mathbf{n}
\end{equation}
\vskip-0.1cm
As the decoder predicts 3D offsets (which we denote by $\mathbf{\hat{y}}$) we use the $L_1(\mathbf{\hat{y}},\mathbf{y})$ loss to train the network for this task.

\vspace*{-0.3cm}
\paragraph*{Ground Truth Error.}
Introduced in Equation 9 of \cite{Gao2020}, $\textrm{GTE}$ describes the sum of squared element-wise differences between the predicted and ground truth plane parameters.
Note that, as some objects in the dataset may have multiple valid symmetry planes, we only compute the $\textrm{GTE}$ for the ground truth plane closest to the predicted one.

\vspace*{-0.35cm}
\paragraph*{Symmetry Distance Error.} 
This metric is used for evaluation and quantifies how closely the original objects match their reflection across the predicted plane.
More precisely, given a set of points $O$ (randomly sampled from an object) and a predicted plane $\hat{\mathcal{S}}$, $\textrm{SDE}$ (as defined in~\cite{Gao2020}) calculates the average squared distance between  $R_{\hat{\mathcal{S}}}(\mathbf{p}), \forall \mathbf{p} \in O$, and its corresponding closest point $\mathbf{q}$ on the object. 

%%%%%%%%%%%%%%%%%%%%%%%%%%%%%%%%%%%%%%%%%%%%%%%%%%%%%%%%%%%%%%%%%%%%%%%%%%%%%%%%%%%%%%%%%%%%%%%%%%%%%%%%%%%%%%%%%%%%%%%%%%%%%%%%%%%%%%%%%%
\section{Experimental Evaluation}
\label{sec:exp}

\begin{table*}[t!]
    \centering
    \footnotesize
    \begin{tabular}{lccccccccc|c}
    \toprule
\makecell[l]{Metric} & 
\makecell{PCA \\\cite{Gao2020}} &
\makecell{Oriented \\ Bounding Box~\cite{Chang2011}}  &
\makecell{Kazhdan \\ \textit{et. al.} \cite{Kazhdan2002}} &
\makecell{Martinet\\ \textit{et. al.} \cite{Martinet2006}} &
\makecell{Mitra\\ \textit{et. al.} \cite{Mitra2006}} &
\makecell{PRST \cite{Podolak2006}\\ with GEDT} &
\makecell{Korman \\\textit{et. al.} \cite{Korman2015}} &
\makecell{PRS-Net\\ \cite{Gao2020}} &
\makecell{Ours} &
\makecell{Ours \\ (Partial)} \\
    \midrule
         GTE ($\times 10^{-2}$) & 2.41 & 1.24 & 0.17 & 13.6 & 52.1 & 3.97 & 19.2 & 0.11 & 7.3 & 19.9 \\
         SDE ($\times 10^{-4}$) & 3.32 & 1.25 & 0.897 & 3.95 & 14.2 & 1.60 & 1.75 & 0.861 & 1.14 & 5.02 \\
    \bottomrule
    \end{tabular}
    \vskip0.1cm
    \caption{
        Performance of our method compared with other methods in literature evaluated using the $\textrm{Ground Truth Error}$ and $\textrm{Symmetry Distance Error}$ metrics on the 1000 objects part of the ShapeNet test set as defined by the PRS-Net paper~\cite{Gao2020}.
        For our method, we show both the results obtained by processing the full 3D objects (directly comparable with the other results in the table), as well as the results we obtain by estimating symmetry planes for partial views of the same objects captured from random viewpoints.
    }
    \label{tab:results_shapenet}
    \vspace*{-0.4cm}
\end{table*}

\begin{table}[t]
    \centering
    \footnotesize
    \begin{tabular}{lcccc}
    \toprule
    ~ & \makecell{mAP} & \makecell{mATE \\ (m)} & \makecell{mASE \\ (1 - IOU)} & \makecell{mAOE \\ (rad)} \\
    \midrule
        \makecell[l]{CenterPoint} & 0.837 & 0.189 & 0.156 & 0.147 \\
        \makecell[l]{CenterPoint + Symmetry} & 0.836 & 0.188 & 0.155 & 0.138 \\
    \bottomrule
    \end{tabular}
    \vskip0.1cm
    \caption{
        Effect of applying the proposed symmetry estimation approach as a refinement step for a 3D object detector.
        We compare the performance of the baseline approach~\cite{Yin2021} \textit{(re-implemented)} with our post-processing on the nuScenes dataset.
        The results show that using the detected symmetries to refine the 3D bounding boxes for \textit{cars} can improve the mAOE metric by more than $6\%$.
    }
    \label{tab:results_nuscenes}
    \vspace*{-0.4cm}
\end{table}

%%%%%%%%%%%%%%%%%%%%%%%%%%%%%%%%%%%%%%%%%%%%%%%%%%%%%%%%%%%%%%%%%%%%%%%%%%%%%%%%%%%%%%%%%%%%%%%%%%%%%%%%%%%%%%%%%%%%%%%%%%%%%%%%%%%%%%%%%%
\subsection{Evaluation on synthetic data}
We experiment on a subset of ShapeNet~\cite{Chang2015} split into training and validation partitions, as in~\cite{Gao2020}, accounting for 80\% and 20\% of the 51,300 models, respectively. 
From the validation partition we hold out the 1000 objects chosen by~\cite{Gao2020}\footnote{\scriptsize{\url{https://github.com/IGLICT/PRS-Net}}} and use them for testing.
We augment the objects by applying random rotations during training and validation.

We train two instances of our architecture: full- and partial-view models.
The former estimates symmetry planes from full 3D objects to compare the performance of our method with other methods in literature~\cite{Kazhdan2002, Martinet2006, Mitra2006, Podolak2006, Chang2011, Korman2015, Gao2020}.  
The latter estimates symmetry planes from objects for which only a partial view is available.
In this case, the partial views are obtained from randomly generated viewpoints (using PyTorch3D's rasteriser~\cite{Ravi2020pytorch3d}).
Both models are trained for the first 200 epochs with the Offsets Loss on full 3D objects then, for 200 more epochs, with both $\textrm{GTE}$ and Offsets Loss on their respective types of objects (full/partial).

As shown in~\autoref{tab:results_shapenet}, our method outperforms several of the competing approaches and reaches an accuracy comparable to the best methods on the \textrm{SDE} metric. 
As for \textrm{GTE}, we attribute the gap between our results and the current state-of-the-art approach to the difference in the methods' designs.
While PRS-Net is devised to predict up to three symmetry planes for each object, our method outputs a single plane as that is enough for the real-world task we are targeting.
For future work, we are planning to investigate an extension of our approach to tackle the multiple symmetry plane scenario.

On the other hand, differently from the methods mentioned above, our approach can estimate reflective symmetry planes from objects for which only a partial pointcloud is available, analogously to the data that would be returned by a sensor in the real world.
We report results on this task in the last column of~\autoref{tab:results_shapenet} and show examples of the estimated planes for both full and partial pointclouds in~\autoref{sec:supplementary}.
The numbers show that the proposed approach can indeed handle partially observed objects; albeit with a slightly lower -- but not significantly different -- performance.
This is encouraging as it shows that our approach can handle objects which are missing significant parts.
Motivated by this finding, we chose to deploy it as part of the real-world pipeline described next.

%%%%%%%%%%%%%%%%%%%%%%%%%%%%%%%%%%%%%%%%%%%%%%%%%%%%%%%%%%%%%%%%%%%%%%%%%%%%%%%%%%%%%%%%%%%%%%%%%%%%%%%%%%%%%%%%%%%%%%%%%%%%%%%%%%%%%%%%%%
\subsection{Evaluation on real data}
To assess the performance of our method on real-world data we use nuScenes~\cite{Holger2019} and focus on the estimation of symmetry planes for vehicles, as they typically have a single symmetry plane running along the length of the object.
We rely on the provided LiDAR sweeps and the ground truth annotated bounding boxes to train the symmetry plane estimator, which is then used to refine the boxes output by one of the top-performing 3D object detectors~\cite{Yin2021}.
Specifically, we update the detections by applying a rigid transform mapping each box to a box having the predicted symmetry plane running along the middle.

\begin{figure}[t]
  \centering
  \includegraphics[width=0.85\linewidth]{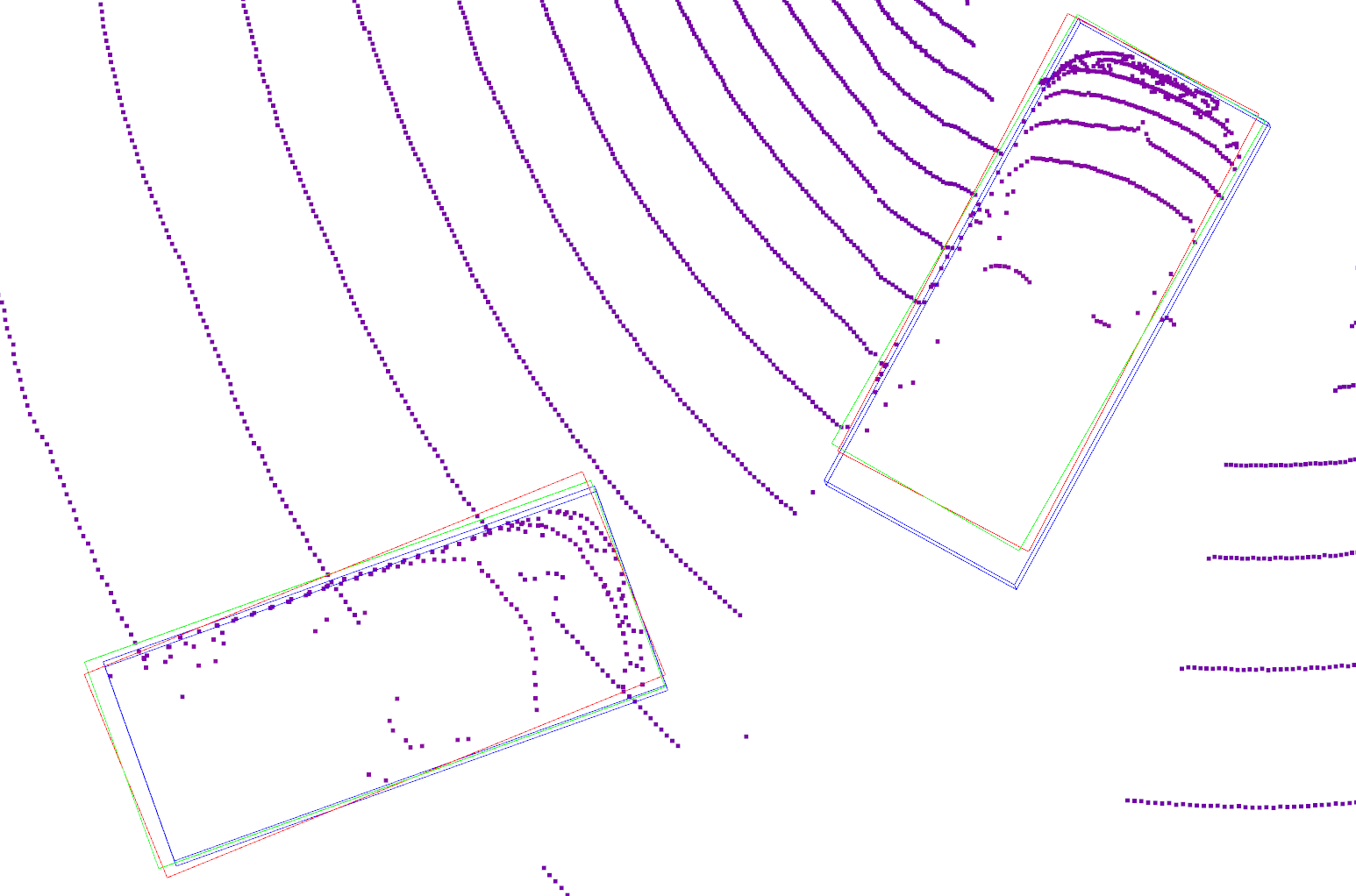}
  \caption{
    A zoomed region of a LiDAR sweep from nuScenes~\cite{Holger2019}.
    In blue we show the ground truth bounding boxes, in red the bounding boxes output by the CenterPoint detector~\cite{Yin2021}, and in green the bounding boxes after refinement.
    Note that the orientation of the refined boxes is closer to the ground truth boxes than the original output of the detector.
    Figure best viewed electronically.
  }
  \label{fig:nuscenes_detections_refinement}
  \vspace{-0.5cm}
\end{figure}

In practical scenarios it is not enough to identify the location of objects surrounding the sensor platform. 
For tasks such as navigation and planning, for example, it is also important to know the accurate heading of all detections as that affects their future pose.
This is especially relevant when both the sensing platform and the detected targets are moving at relatively high speeds, as small orientation errors can contribute to large errors in the forecasted state of the system.
\autoref{fig:nuscenes_detections_refinement} illustrates the effects of using the estimated symmetry plane to update vehicle detections.
Quantitatively,~\autoref{tab:results_nuscenes} compares the symmetry refinement to the raw output of the baseline detector using the nuScenes evaluation metrics.
The main translation and scale metrics are not affected by this post-processing -- which is expected, as symmetry plane refinement mainly affects the orientation of each box.
On the other hand, the Mean Average Orientation Error for objects of the \textit{cars} category is improved by more than $6\%$, thus allowing for better decision making in downstream tasks.

%%%%%%%%%%%%%%%%%%%%%%%%%%%%%%%%%%%%%%%%%%%%%%%%%%%%%%%%%%%%%%%%%%%%%%%%%%%%%%%%%%%%%%%%%%%%%%%%%%%%%%%%%%%%%%%%%%%%%%%%%%%%%%%%%%%%%%%%%%
\section{Conclusion}
In this paper we presented a method that can be used to estimate planar reflective symmetry planes from 3D objects or their partial views, captured from arbitrary viewpoints.
The latter is important because, in real-life scenarios, sensors mounted on robotic platforms or autonomous vehicles are unable to observe the full extents of target objects, as they are limited to capturing information from a single (visible) side. 
Estimating the location of symmetry planes from only a view of an object can help determine its extents and reconstruct an approximation of the full 3D object model by reflecting the visible points across the detected plane.

We showed that the proposed approach can be deployed satisfactorily in a real-world 3D object detection pipeline as a post-processing step refining the 3D bounding boxes to increase the detection accuracy.
This can help by improving the performance of several subsequent tasks -- such as segmentation, navigation, planning, grasping or trajectory forecasting -- which are typically part of robotics or autonomous driving systems.

%%%%%%%%%%%%%%%%%%%%%%%%%%%%%%%%%%%%%%%%%%%%%%%%%%%%%%%%%%%%%%%%%%%%%%%%%%%%%%%%%%%%%%%%%%%%%%%%%%%%%%%%%%%%%%%%%%%%%%%%%%%%%%%%%%%%%%%%%%
\section*{Acknowledgements.}
\noindent The authors wish to thank Dr. Stuart Golodetz for insightful discussions and helpful pointers.

{\small
\bibliographystyle{ieee_fullname}
\bibliography{bibliography}
}

%%%%%%%%%%%%%%%%%%%%%%%%%%%%%%%%%%%%%%%%%%%%%%%%%%%%%%%%%%%%%%%%%%%%%%%%%%%%%%%%%%%%%%%%%%%%%%%%%%%%%%%%%%%%%%%%%%%%%%%%%%%%%%%%%%%%%%%%%%
\newpage
\appendix
\section{Qualitative Results}
\label{sec:supplementary}
\begin{figure*}[t]
  \centering
  \includegraphics[width=0.945\linewidth]{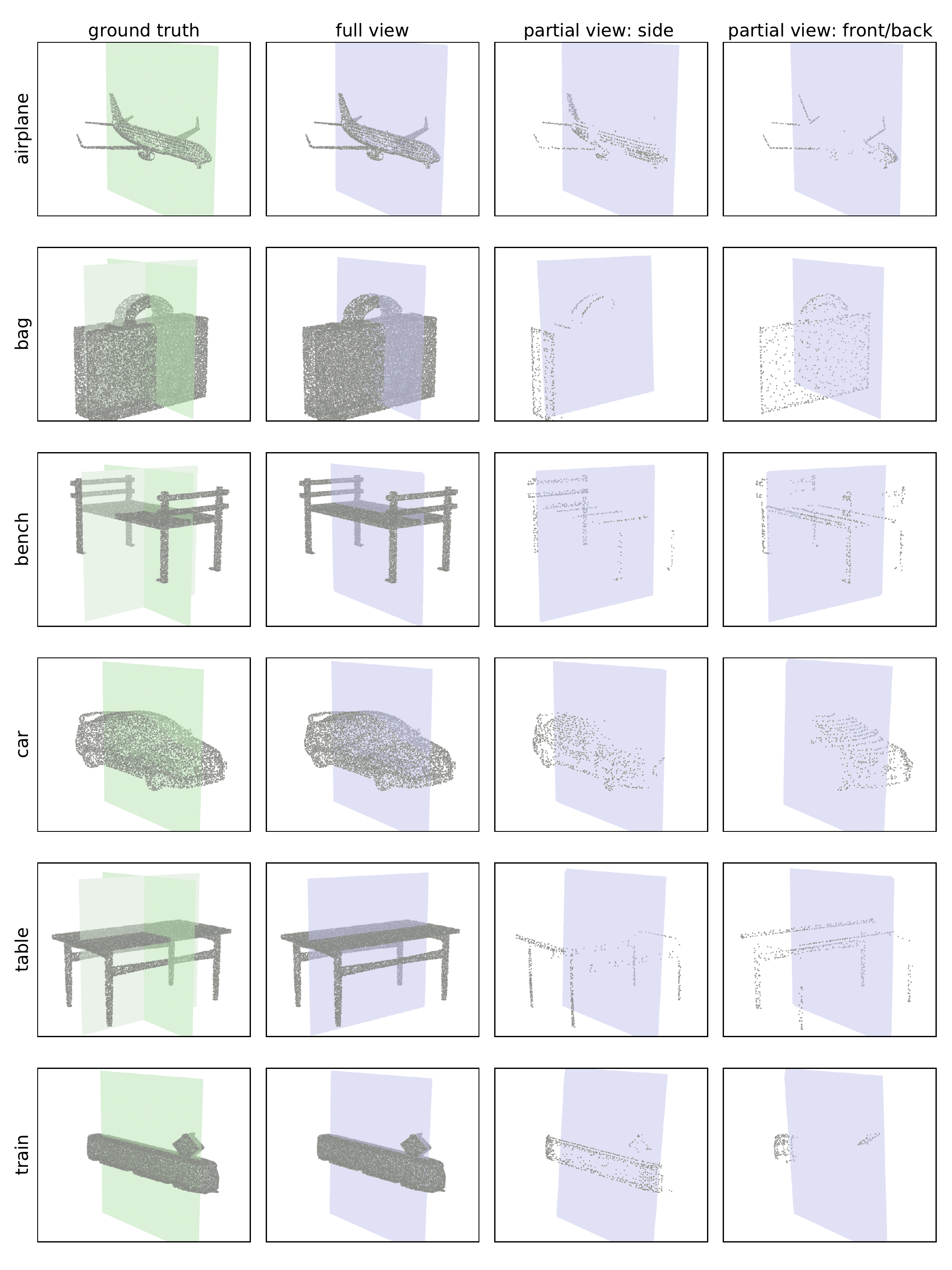}
  \caption{
    Examples of the estimated planes for both full and partial pointclouds from ShapeNet~\cite{Chang2015}.
  }
  \label{fig:supplementary}
\end{figure*}

In \autoref{fig:supplementary}, we show examples of symmetry planes estimated by our approach from both full and partial pointclouds from ShapeNet~\cite{Chang2015}. 
The first column shows the object models together with their corresponding ground truth symmetry planes (in green).
Note that, for objects that have multiple reflective symmetries, we show all ground truth  planes.

The following columns illustrate the planes estimated by our model in different scenarios.
The second column shows the planes (in blue) output by the model when processing full pointclouds.
The last two columns correspond to the planes estimated by the model when processing partial pointclouds: either observed from one lateral side  or from a front (or back) view.

It is worthwhile noting that the pointclouds that our method can handle are missing significant parts.
Our approach currently outputs a single symmetry plane per object. 
If the object has multiple valid symmetry planes, depending on the viewpoint used to rasterise the partial pointcloud, the predicted plane might differ. 
For example, as in the \textit{bag} row, the estimated plane output for the side viewpoint is orthogonal to the plane predicted for the back view -- in spite of this, both predictions are valid reflective symmetry planes. 

\clearpage
\thispagestyle{empty} % Disable page number since it shows on top of the caption.

\end{document}